\RequirePackage{fix-cm}
\documentclass[smallextended]{svjour3}       
\smartqed  
\usepackage{times}
\usepackage{epsfig}
\usepackage{graphicx}
\usepackage{amsmath}
\usepackage{amssymb}

\usepackage{soul}
\usepackage{url}
\usepackage[utf8]{inputenc}
\usepackage{booktabs}
\usepackage{multirow}
\usepackage{blindtext}
\usepackage{tabularx}
\usepackage[table,xcdraw,dvipsnames]{xcolor}
\usepackage[]{xcolor}
\usepackage{todonotes}
\usepackage{array} 
\usepackage{color}
\usepackage{xspace}
\usepackage{fancyhdr}
\usepackage{csquotes}
\usepackage{graphics}
\usepackage[pagebackref=true,breaklinks=true,colorlinks,bookmarks=false]{hyperref}
\usepackage[T1]{fontenc}
\usepackage{pifont}

\usepackage{tikz}
\usepackage{pgfplots}
\usetikzlibrary{patterns}
\pgfplotsset{compat=1.16}
\usepackage{wrapfig}
\usepackage{subcaption}
\usepackage{smartdiagram}  
\usepackage[edges]{forest}
\usetikzlibrary{arrows.meta}
\usepackage{rotating}

\smartdiagramset{
uniform color list=blue!20 for 8 items, 
bubble center node color = blue!40,
}

\newcommand{\latinphrase}[1]{\textit{#1}}
\newcommand{\etal}{\latinphrase{et~al.}\xspace}
\newcommand{\ie}{\latinphrase{i.e.}\xspace}

\newcommand{\etc}{\latinphrase{etc.}\xspace}

\usepackage{tikz}
\def\checkmark{\tikz\fill[scale=0.3](0,.35) -- (.25,0) -- (1,.7) -- (.25,.15) -- cycle;}
\begin{document}

\title{{An Overview of Violence Detection Techniques: Current Challenges and Future Directions}
}


\author{Nadia Mumtaz        $^\dag$\and
        Naveed Ejaz      $^\dag$\and
        Shabana Habib    \and
        Syed Muhammad Mohsin \and
        Prayag Tiwari \and
        Shahab S. Band $^\dag$\and
        Neeraj Kumar $^\dag$
}


\institute{N. Mumtaz, N. Ejaz\at
            Department of Computer Science, Iqra University Islamabad Campus, Pakistan \\ 
            \email{nadiamumtaz25763@iqraisb.edu.pk, naveed.ejaz@iqraisb.edu.pk} 
        \and
            S. Habib \at 
                Department of Information Technology, College of Computer, Qassim University, Buraydah 52571, Saudi Arabia\\
                \email{s.habibullah@qu.edu.sa} 
        \and
            S. M. Mohsin \at
            Department of Computer Science, COMSATS University Islamabad, 45550, Pakistan\\
            \email{syedmmohsin9@yahoo.com} 
        \and 
            P. Tiwari \at
            School of Information Technology, Halmstad University, Sweden\\
            \email{prayag.tiwari@hh.se}
        \and 
            Shahab S. Band \at
            Future Technology Research Center, National Yunlin University of Science and Technology, 123 University Road, Section 3, Douliou, Yunlin 64002, Taiwan \\
            \email{shamshirbands@yuntech.edu.tw}
        \and 
            N. Kumar \at
            Department of Computer Science and Engineering, Thapar Institute of Engineering and Technology (Deemed University), Patiala (Punjab), India\\
            \email{neeraj.kumar@thapar.edu} 
        \and 
            $^\dag$ Corresponding authors: Nadia Mumtaz, Naveed Ejaz, Shahab Shamshirband, Neeraj Kumar.\\
              Tel.: +91-8872540189\\
         }

\date{Received: date / Accepted: date}

\maketitle

\begin{abstract}
The Big Video Data generated in today's smart cities has raised concerns from its purposeful usage perspective, where surveillance cameras, among many others are the most prominent resources to contribute to the huge volumes of data, making its automated analysis a difficult task in terms of computation and preciseness. Violence Detection (VD), broadly plunging under Action and Activity recognition domain, is used to analyze Big Video data for anomalous actions incurred due to humans. The VD literature is traditionally based on manually engineered features, though advancements to deep learning based standalone models are developed for real-time VD analysis. This paper focuses on overview of deep sequence learning approaches along with localization strategies of the detected violence. This overview  also dives into the initial image processing and machine learning-based VD literature and their possible advantages such as efficiency against the current complex models. Furthermore,the datasets are discussed, to provide an analysis of the current models, explaining their pros and cons with future directions in VD domain derived from an in-depth analysis of the previous methods. 

\end{abstract}

\section{Introduction}
\begin{figure}
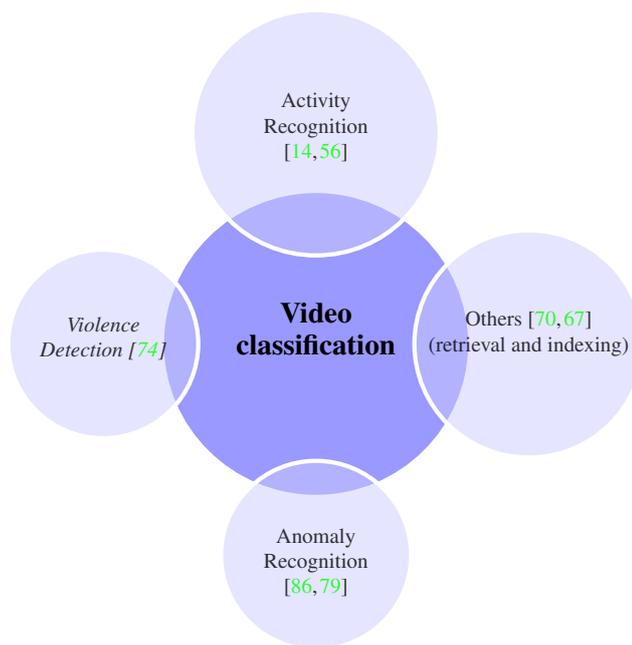

    \centering
    
    
    \smartdiagram[bubble diagram]
    { \textbf{Video} \\ \textbf{classification} \\, \\ \\ \\ Activity \\ Recognition \\ \cite{dang2020sensor,pareek2021survey} \\ \\ \\,
    \\ \textit{Violence}\\\textit{Detection \cite{sumon2020violence}} \\, \\ \\ Anomaly \\ Recognition \\ \cite{ullah2021efficient,ullah2020one} \\ , Others \cite{spolaor2020systematic,shen2020advance} \\ (retrieval and indexing) \\}

    \caption{Video classification categorization into sub-groups. The diameter of the circle indicates the maturity level of the field which is selected based on its reputation, number of datasets, and endorsement of related research community in terms of citations. The italic text refers to the currently observed topic. }
    \label{fig:videoclassificationdomains}

\end{figure}

In today's modern world of 24/7 surveillance, vision sensory data \textcolor{black}{are} widely used to monitor activities automatically and report them to connected departments for counter actions. Automated surveillance is a major problem and concern of computer vision experts, where deep models have achieved tremendously precise results in many computer vision problems, ranging from object detection to complex multiple activities prediction and perception. The achievements of computer vision techniques are mainly due to the excessive demand of automated video analytic methods for enormous applications such as video indexing and retrieval, video summarization~\cite{wu2020dynamic}, action and activity recognition \cite{dang2020sensor}, and video classification~\cite{sasithradevi2020video}. In video classification domain, action and activity recognition has received significant attention in research community due to its applications to medical healthcare, security, sports and entertainment, among many others. Where violence detection (VD) is among other video classification major groups focused on detection of harmful and brutal events patterns from an input video. Broader-level categorization of video classification domains is given in Figure~\ref{fig:videoclassificationdomains}.

VD refers to abnormal patterns selection$/$extraction from a normal surveillance video that could occur for a very short or a prolonged time duration. Its early detection assists in either prevention or reducing the residual damage in terms of human lives and their properties by initiating countermeasures such as notifying the nearest concerned departments for responsive actions. Violence can be any abnormal brutal action by humans including hitting someone, damaging, destroying, and injuring.

\begin{figure}
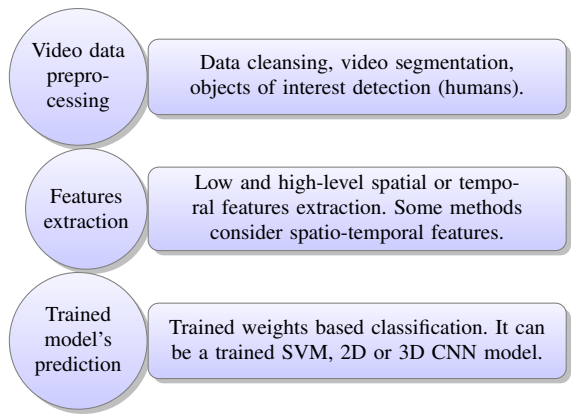

    \centering
    
    \smartdiagram[descriptive diagram]
    {  
    {Video data preprocessing, {Data cleansing, video segmentation, objects of interest detection (humans).}},
    {Features extraction,{Low and high-level spatial or temporal features extraction. Some methods consider spatio-temporal features.}},
    {Trained model's prediction,{Trained weights based classification. It can be a trained SVM, 2D or 3D CNN model.}},
    }
    
    \caption{Major steps involved in VD. }
    \label{fig:VDsteps}
    
\end{figure}

\textcolor{black}{Video data (VD) based on video data is applicable to a large number of real-world scenarios. It is evident that the occurrence of violence is extremely rare when compared to the normal pattern of continuous video surveillance. As a result, deploying human resources to monitor video streams could be impractical, as it would require repetitive practice in order to detect abnormal patterns. Currently, automated video detection techniques are of great assistance in monitoring surveillance video streams in real-time. The main goal of a VD system is to detect and report any abnormal behavior, which is deviated from the normal pattern of behavior. Therefore, instead of human intervention, VD can be used effectively in real-world surveillance scenarios. \textcolor{black}{Based on the categorized video data collected from CCTV cameras, the VD techniques utilise different soft computing mechanisms to learn the patterns and analyze the types of motions associated with normal and abnormal behaviors ~\cite{chang2022hybrid} in order to detect violent and non-violent situations using the collected data ~\cite{mumtaz2022fast}}} VD using \textcolor{black}{soft computing techniques} techniques from video data is in practice since decades \cite{datta2002person}, initiated based on traditional image processing techniques. The early VD literature has consideration of various attributes for decision making such as acceleration of human motion, their appearance, and motion flow, among many others. The generic steps involved in baseline research include video data preprocessing, features extraction, and its final classification into violent or non-violent segments. The preprocessing step involves data cleansing and video segmentation. Features extraction refers to individual-level (spatial) frames processing using certain features extractors such as histogram oriented gradients (HOG), motion, speed, and optical flow. The final classification of these features is based on an existing model's weights and configuration acquired using the training set of the dataset. The models used in pioneering VD research are machine learning based techniques such as support vector machine (SVM). These steps are given in Figure~\ref{fig:VDsteps}.

With the advent and advancements of deep learning in computer vision domains, researchers also employed it for VD. The basic pipeline of deep learning based VD is closely similar to the traditional mechanism, but the absence of preprocessing in most of the deep learning based strategies is a highlighted distinction. In most of the cases, a deep VD model has an end-to-end architecture to input a video (sequence of frames) or a single frame and output their probability of being violent or normal.

While there has been extensive research in activity recognition domain \cite{dang2020sensor}, VD, being its sub-domain has not been explored to its value in real-world problems. Although abnormal activity recognition too has been studied widely, but that is different from violent activity recognition. There exist several challenges in VD domain, the foremost of which is the limited datasets publicly available. Real-time performance with significant accuracy to be implemented in a smart city's surveillance scenarios has yet to be achieved. Embedded devices with intelligent vision have been recently applied to many emerging computer vision domains, need to be applied in VD domain. 

\begin{figure*}[!h]
    \centering
    \includegraphics[width=\textwidth]{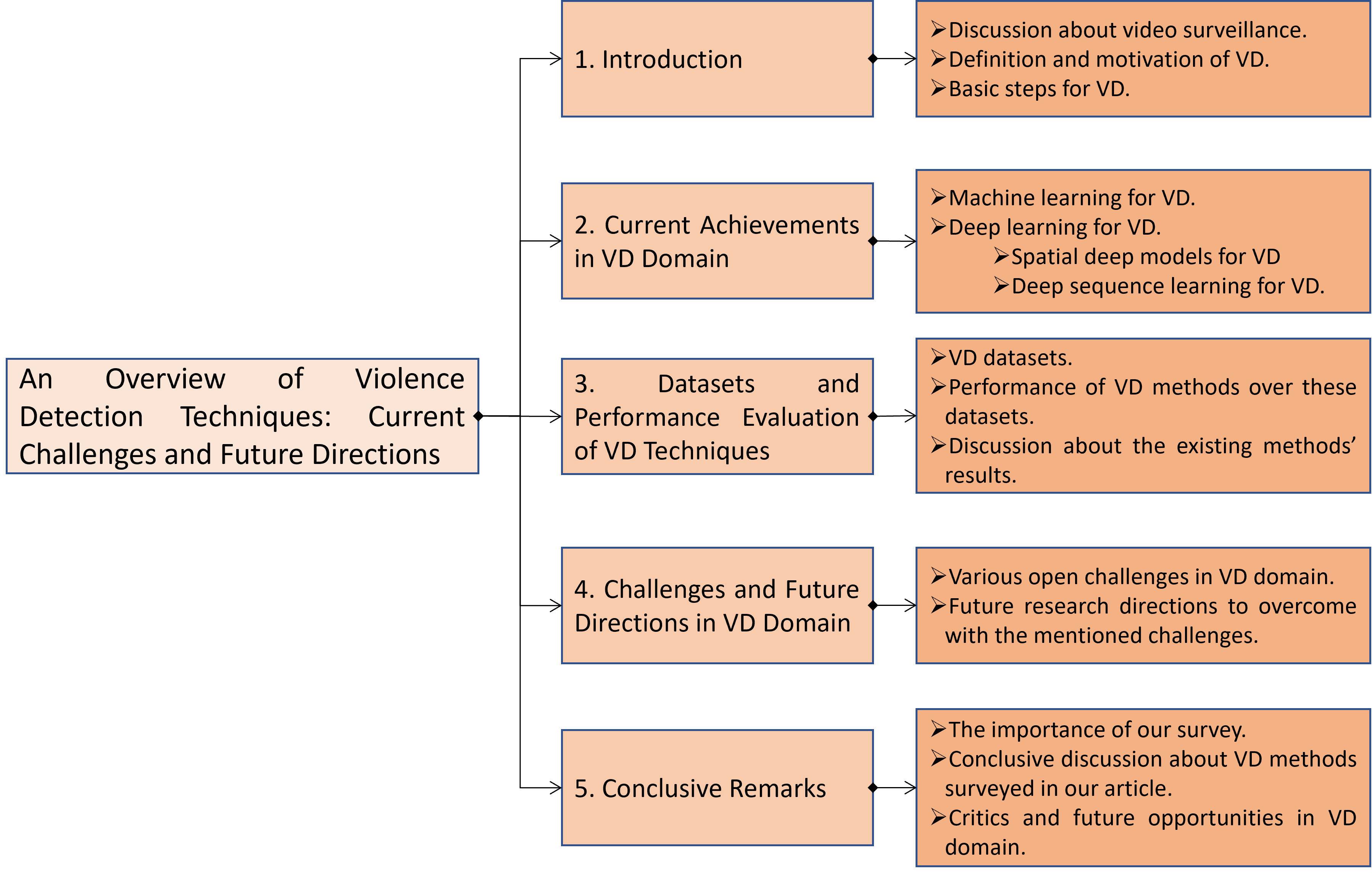}
    \caption{The overall structure of the proposed survey.}
    \label{fig:structureofsurvey}
\end{figure*}

The mentioned challenges and their further extension along with future derivations based on the current literature are provided in this review article. The proposed survey flows smoothly from traditional machine learning based techniques towards the recent deep learning based models. Further, we explain basic steps and conclude the generic flow of VD methods, supported by baseline references for further guidance of readers. Our summarized contributions can be enumerated as follows:

\begin{enumerate}
    \item Detailed analysis of existing VD literature, discussion on representative methods from classified VD categories, as given in Figure~\ref{fig:structureofsurvey}.
    \item Explanation of advantages and drawbacks of various VD methods based on their utilized features.
    \item Performance evaluation of representative VD methods \textcolor{black}{and discussion about how and why some methods achieved higher accuracy with fewer parameters.}
    \item Comprehensive future insights for interested readers and domain experts. The future directions are derived based on the analyzed literature and supported by appropriate references.
\end{enumerate}

\textcolor{black}{The VD techniques that broadly falls under the scope of video classification are introduced in this section. The major contributions of our survey are highlighted in bulleted form.} The rest of the paper has 4 sections. Section 2 corresponds to the detailed explanation of the current achievements in VD domain. In Section 3, we best explain the available datasets and discuss the performance evaluation of different methods on these datasets. In Section 4, we highlight the current challenges in VD domain, each challenge followed by its corresponding future research guidelines. Finally, we conclude our article in Section 5 by explaining the derivation of this research.

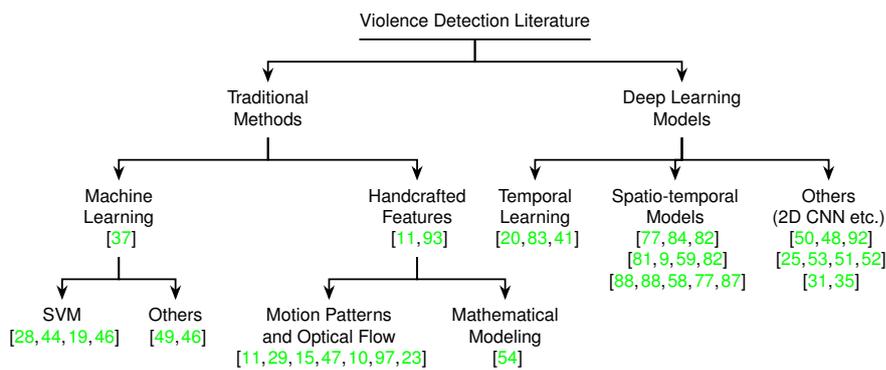
\begin{figure*}
    \centering
\resizebox{\columnwidth}{!}{
\begin{forest}
  for tree={
    align=center,
    parent anchor=south,
    child anchor=north,
    font=\sffamily,
    edge={thick, -{Stealth[]}},
    l sep+=10pt,
    edge path={
      \noexpand\path [draw, \forestoption{edge}] (!u.parent anchor) -- +(0,-10pt) -| (.child anchor)\forestoption{edge label};
    },
    if level=0{
      inner xsep=0pt,
      tikz={\draw [thick] (.south east) -- (.south west);}
    }{}
  }
  [Violence Detection Literature
    [Traditional\\Methods
      [Machine\\Learning\\~\cite{hussain2020multi}
      [SVM\\~\cite{gao2016violence,lohithashva2020violent,dhiman2017high,mabrouk2017spatio}]
      [Others\\~\cite{mishra2018automated,mabrouk2017spatio}]]
      [Handcrafted\\Features\\~\cite{chen2011violence,xu2014violent}
        [Motion Patterns\\and Optical Flow\\~\cite{chen2011violence,gracia2015fast,datta2002person,mahadevan2010anomaly,chen2008recognition,zhang2016new,febin2020violence}]
        [Mathematical\\Modeling\\~\cite{nguyen2005learning}]
      ]
    ]
    [Deep Learning\\Models
      [Temporal\\Learning\\~\cite{ding2014violence,ullah2019violence,li2019efficient}]
      [Spatio-temporal\\Models\\~\cite{traore2020violence,ullah2020cnn,ullah2021intelligent}\\~\cite{ullah2021ai,chang2022hybrid,qiu2017learning,ullah2021intelligent}\\~\cite{vosta2022cnn,vosta2022cnn,peixoto2019toward,traore2020violence,vijeikis2022efficient}\\]
      [Others\\(2D CNN etc.)\\~\cite{mu2016violent,meng2017trajectory,xia2018real}\\~\cite{freire2022inflated,naik2022automated,mumtaz2022fast,mumtaz2018violence}\\~\cite{hanson2018bidirectional,hussain2022real}]
      ]
    ]
  ]
\end{forest}
}
\caption{A comprehensive form of overall VD literature.}
\label{fig:VDLiterature}
\end{figure*}

\begin{figure*}[!h]
    \centering
    \includegraphics[width=\textwidth]{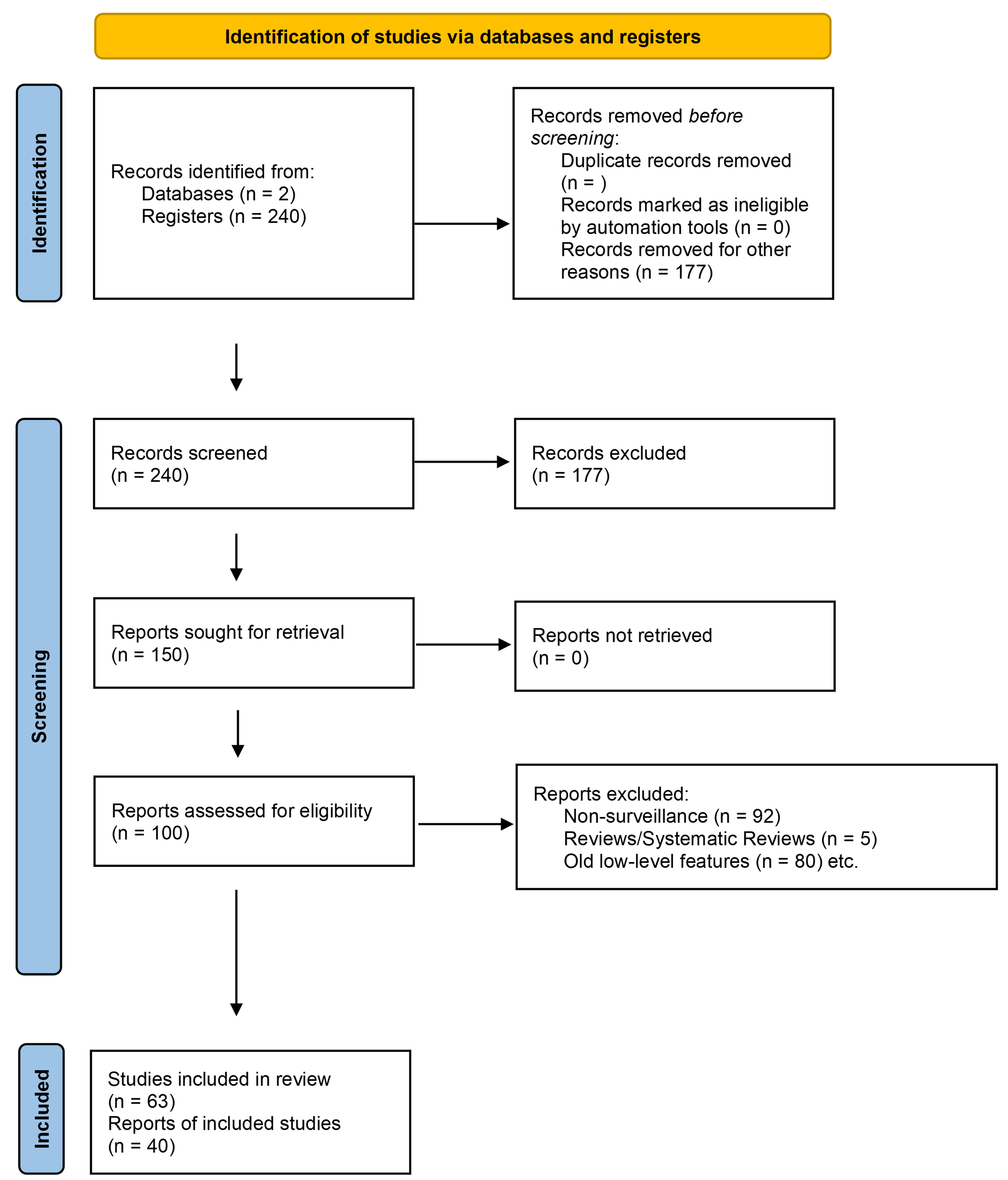}
    \caption{The overall process of research articles retrieval from different databases.}
    \label{fig:prismamodel}
\end{figure*}

\textcolor{black}{
\section{Review Methodology}
The VD domain using video data for surveillance applications is in action since 2002, when Datta~\etal~\cite{datta2002person} presented their first person-to-person VD approach. Afterwards, in the era between 2002 and 2013, mainstream techniques are based handcrafted features and completely rely on the domain knowledge to constitute low-level features from the input videos. With the advancements in deep learning models, the very first approach utilizing 3D ConvNets was introduced by Ding~\etal~\cite{ding2014violence}, that works independently without any prior knowledge. Following VD using 3D ConvNets, many researchers introduced deep learning-based techniques, that show significantly better results when compared to the hand-engineered features methods. }

\textcolor{black}{
In this review, the baseline research contributions in VD domain are discussed, leading to the current state-of-the-art approaches and advanced deep models. The research articles are retrieved from several famous databases such as \emph{Web of Science} and \emph{Google Scholar}. The overall PRISMA flow diagram for the research articles retrieval process is visualized in Figure~\ref{fig:prismamodel}. Overall there are lots of articles retrieved during the initial search, that are first scrutinized using the title. For instance, the research contributions mentioning non-surveillance domains violence are not included in our review. Similarly, lots of low-level features-based articles are available in the early VD literature, but the ones with comparatively higher number of citations are discussed in the proposed survey.}

\section{Current Achievements in VD Domain} \label{sec:Relatedworks}
The growth and technological advancements in the field of computer vision has widely impacted the surveillance applications~\cite{dehingia2022help} positively, providing various services such as continuous video data monitoring using VD and reporting. There are a lot of existing techniques focusing on these services, that are broadly categorized into handcrafted (conventionally known as low-level features and machine learning) and deep learning-based methods, discussed separately below and given hierarchically in Figure~\ref{fig:VDLiterature}. \textcolor{black}{
The manuscript retrieves contributions based on popularity amongst computer vision and visual perception experts. Each piece of research is reviewed for relevance after retrieval from renowned research repositories such as \emph{Web of Science}, where we used different keywords such as \emph{violence detection in video data, video violence detection, violence detection,~\etc}. The literature review in this proposal focuses on papers that have been published between 2002 and the present date. It is notable that deep learning results are more prevalent during the retrieval procedure. Finally, the relevance of each research is checked~\ie using the mentioned technique's strategy, input horizon, the intermediate procedure, and finally the output results. For instance, research articles focused on VD using sensors other than vision are not considered in the proposed survey. Similarly, hybrid techniques considering various sensory data are also excluded from discussion in this research.}

\begin{table*}[ht]
    \centering
    \caption{Traditional ML representative VD methods selected based on their popularity among VD experts.}
    \resizebox{\textwidth}{!}{
    \begin{tabular}{m{1.5cm}||m{4.5cm}m{3cm}m{3cm}m{3cm}}
    \toprule
    \textbf{Ref.} & \textbf{Strategy} & \textbf{Horizon} & \textbf{Dataset$/$Availability} & \textbf{Remarks}
    \\ \midrule
    Chen \etal \cite{chen2011violence} & Violence classification using SVM & Movies data & Movies used in experimentation$/\checkmark$ & In traditional machine learning mechanisms, mainstream researchers utilize the potentials of SVM to detect violence in surveillance videos. \\ \midrule
    
    Hassner \etal \cite{hassner2012violent} & VIolent Flows (ViF) descriptor are used to classify violent or nonviolent using linear SVM & Surveillance & Action Similarity Labeling Challenge (ASLAN) and Hockey dataset$/\checkmark$ & This is a challenging dataset against other datasets due to data diversity and  reduced amount of videos present in this dataset. \\ \midrule
    
    Deniz \etal \cite{deniz2014fast}  & Spatio-temporal features and acceleration computation for VD & Surveillance and sports & UCF101, Hockey Fight, and Movies data$/\checkmark$ & Fast VD method \\ \midrule
    
    Gracia \etal \cite{gracia2015fast}  & Motion blob and acceleration measure vector method is used for fight detection & Surveillance crowd & UCF-101, Hockey Fight, and Movies$/\checkmark$ & Fast computation over the cost of accuracy. \\ \midrule
    
    Gao \etal \cite{gao2016violence} & Motion information with AdaBoost algorithm for features selection and linear SVM for classification & Surveillance & Crowd violence and hockey fight$/\checkmark$ & Hybrid motion features with AdaBoost and Linear SVM performs well against SOTA.   \\ \midrule
    
    Bilinski \etal \cite{bilinski2016human} & Proposed improved fisher vector to recognized violence in videos & Surveillance & Violent-Flows dataset, Hockey Fight dataset and Movies dataset$/\checkmark$ & Improved efficiency with violence detection and recognition potentials. \\ \midrule
    
    Li \etal \cite{li2016detecting} & Proposed Multi-modal features model by exploiting subclass visualization to manually labeling violence in videos & Surveillance & MediaEval 2015 violence dataset$/\checkmark$ & Diverse set of sub-classes video annotation and classification over MediaEval 2015 dataset \ie extending this dataset. \\ \midrule
    
    Dhiman \etal \cite{dhiman2017high}  & Proposed integrated features-based on average energy images framework to recognize abnormal activities & Surveillance & R fall detection and KARD dataset$/\times$ & Hybrid features from Kinect and visual sensor are reduced using PCA and classified via SVM.  \\ \midrule
    
    Malek \etal \cite{al2017novel}  & Temporal differencing algorithm and motion regions are located using gaussian function and SVM for normal and abnormal activities recognition & Surveillance & Private students recorded data$ /\times$& After VD, retrieval of the detected objects from database for detailed analysis. \\ \midrule
    
    Mishra \etal \cite{mishra2018automated}  & Bag of words representation technique is used for detection and classification of art material videos  & Surveillance & KTH$/\checkmark$ and proposed Judo and Taekwondo$/\times$ & Simple martial arts videos classification using statistical machine learning.  \\ \midrule
    
    Song \etal \cite{song2018multi}  & Single temporal and multi-frameworks based on learning or semantic modeling technique for high level activity analysis & Surveillance & BEHAVE dataset $/\times$, \newline YouTube data $/\checkmark$, and \newline NUS–HGA dataset$/\checkmark$ & Graph trees are used in the framework and testing on real-world YouTube videos show promising generalization abilities of this method.  \\ \midrule
    
    Mabrouk \etal \cite{mabrouk2017spatio} & Crowd$/$non-crowd violence detection using covariance matrix and spatio-temporal features. Optical flow of HOG features & Crowded $/$ Non-crowded areas VD & Hockey fight and violent flows dataset $/\checkmark$ &  In this method an SVM classifier is trained to classify diverse data, yielding poorly generalized model, showing 88.6 $/$ \% accuracy for hockey fight dataset \\ \midrule
    
    Lohithashva \etal \cite{lohithashva2020violent} & LBP and GLCM features with supervised classifiers for VD & Surveillance and movies & Hockey fight and violent-flow $/\checkmark$  & Low-level features show reduced performance and limited generalization abilities. \\ \midrule
    \end{tabular}
    }
    \label{tab:MLforVD}
\end{table*}
\subsection{Machine Learning Techniques for VD}
These are baseline and early VD methods~\cite{de2010violence}, considering purely image processing and pattern recognition methods to predict violence and abnormal patterns using either filtering techniques motion information. This group of study is effective for simple VD problems with only abnormal motion patterns, where limited to work in situation with instant normal motion and complex real-world situation. In low-level features-based approaches, authors generally use human trajectories and motion information from visual data \cite{chen2011violence}. In motion information, most of the methods use optical flow or its several invariants. In supervised learning-based traditional mechanisms, mainstream researchers utilize the potentials of Support Vector Machine (SVM). Some of the research in VD domain is based on multi-modalities such as aural and video data are employed to detect violence in surveillance videos. In the subsequent paragraphs, these methods are discussed briefly. 

The human violence refers to fist fighting, hitting other objects, and many other similar actions that occur in surveillance videos. To detect such activities automatically, a person-on-person violence detection method is introduced in \cite{datta2002person}, that is functional for video data. The authors utilized motion trajectories information along with the limb’s orientations information in their method to detect violence effectively. There is an acceleration measure vector in their method which contains information about the magnitude of motion and jerk is defined here using temporal derivative of the acceleration motion vector. This method is completely based on low-level features \ie background subtraction and acceleration measure vector computation. 

Continuing research using low-level features, Nguyen \etal \cite{nguyen2005learning} used hidden Markov model for activities recognition that may contain violence. Their main novelty and contribution lie in construction of a particle filter with an approximation inference scheme that computes the distributions of hidden Markov model at constant time to receive new observation and data, making the system functional in real-time. 

Heading onward towards more sophisticated and reliable anomaly and violence detection methods \cite{mahadevan2010anomaly}, Mahadevan \etal introduced a novel framework for anomaly detection in crowded scenes~\cite{naik2022automated}, that is a very challenging task to achieve. Their research concludes that a set of three properties acts as the main constituent of any anomaly detection system. These properties include integrated modeling of dynamics and appearance of the ongoing scene in an input video, temporal, and spatial abnormalities. Their model is completely based on phenomenon that the normal behavior has a specific pattern and contains a mixture of predicted, but at the same time dynamic texture. While detecting any outlier in this pattern or dynamics indicates anomaly. 

In continuation with handcrafted features-based methods, motion vectors are widely used as well in the early related literature. As Chen \etal utilized human behavior using binary location motion patters and decide about an aggressive behavior \cite{chen2008recognition}. Similarly, optical flow-based methods perform comparatively better. For instance, in a baseline research \cite{zhang2016new}, the authors used Gaussian Model of Optical Flow to effectively extract candidate violence regions, that are adaptively modeled as a deviation from any normal behavior being observed in the current scene. These candidate violence regions are then sampled to generate or detect violence.

A deep study of the aforementioned methods reveals their several drawbacks, which limits their adaptation and deployment in real-world surveillance scenarios. Most of these methods are designed to function for static image data, that is the foremost concern of today’s smart surveillance, where actions and events occur in a sequence of images. Therefore, these methods do not result in trusty-worthy decisions due to their spatial image processing. Another major concern related to these methods is their features representation approaches, which are very confined and are extremely low-level to be trusted for smart surveillance scenarios. In advanced machine learning based methods, the SVM~\etc perform well for classifications and activity recognition tasks~\cite{hussain2020multi}, but when the features are extracted using handcrafted techniques, the output results are poorly generated. The traditional methods are comprehensively tabulated in Table \ref{tab:MLforVD}, where we also indicate the used dataset of each method. To overcome the limitations mentioned above, deep learning performs very well, and deep learning-based approaches are discussed in the subsequent section.

\subsection{Deep Learning for VD}

The emergence of deep learning models in several computer vision domains \textcolor{black}{motivated researchers to apply it} for enormous surveillance applications including violence detection etc. Video analysis is a challenging problem due to the involvement of complex image data and its huge processing for useful information extraction. \textcolor{black}{Video analysis comprises several major domains, where primarily it is composed of video classification~\cite{wang2018appearance,tran2019video,li2020smallbignet}, segmentation~\cite{gao2022deep,wang2021survey}, among many others. Video classification~\cite{yue2015beyond} is an interesting domain with main focus on effective temporal contents representation that significantly contributes to precise video label prediction. Although the early approaches in video classification are based on simple CNNs~\cite{karpathy2014large}, but the recent methods employ various temporal~\cite{carreira2017quo} and spatio-temporal~\cite{hussain2022vision} strategies for video classification. Video classification is further divided into several major domains such as activity recognition, anomaly detection and recognition, and violence detection and recognition.}

Violence occurs in sequence of frames, falling under the broad domain of \textcolor{black}{video} classification. Sequential frames data, compared to single image data is much more complex due to involvement of time series analysis and its patterns and structure learning for final output generation. Motion features are also used in sequential form such as in bag of words, as presented in an existing research~\cite{fu2017automatic}. The main strategy used in this system is to analyze the motion to detection fight events. Although, motion features involve sequential patterns processing, but still motion features seem to perform poorly, as huge motion in sequential frames do not necessarily guarantee abnormal events i.e., it can be normal patterns. Otherwise, anomaly or violence may occur with slight motion, therefore limiting its usage.

Towards violence detection, deep learning-based methods~\cite{roman2020violence} have completely replaced the traditional working of hand-crafted features-based algorithms with convincingly precise output results. As mentioned earlier, violence occurs in sequence of frames, therefore, many researchers use spatial as well as temporal features to generate optimal results. Spatial features are concerned with the frames-level data, while temporal features works best for sequence analysis. In activity classification domain, the concept of hybrid features, i.e., integrating spatial and temporal features is very common. For instance, Amira Ben and Zagrouba~\cite{mabrouk2017spatio} employed spatio-temporal optical flow in their research to detect violence. Their method estimates the magnitude and orientation of optical flow, followed by its descriptor and both these (estimator and descriptor) are used to train an SVM classifier, showing 88.6\% accuracy for hockey fight dataset. This method is based on traditional machine learning, yielding poorly generalized model, as SVM is not a good option to classify diverse data.
In a follow up research~\cite{ullah2019violence}, C3D features are used to train a deep learning model for violence detection. Deep learning features are used as intermediate representation of sequential information, followed by SoftMax classifier to extract the probability of the class having maximum chances of occurrence. This research shows higher computational complexity and is not real-time, as reported by authors in their experiments as 25 FPS. In another recent research, authors used deep learning features with long short-term memory (LSTM) for anomaly detection~\cite{ullah2020cnn}. The spatio features representation in this method is achieved using the famous ResNet-50 model, where the authors extracted already learned features from ResNet-50 pretrained model in their method. LSTM is used in this method for sequential learning, where the learned features are stacked together to form a sequence of 15 frames, that are passed to bi-directional LSTM for anomalous activities patterns learning.

The aforementioned methods tend to tackle the challenges of hand crafted-features based violence detection systems, where they perform much better to classify violent activities. But the time complexity of these models restrict their practical deployment in real-life surveillance scenarios, such as implementation in smart cities surveillance etc. In addition, mainstream deep models provide only violence detection or classification services, demanding human experts for its analysis and reporting ultimately to the concerned departments.

\begin{table*}
    \centering
    \caption{Deep learning representative VD methods selected based on their popularity among VD experts.}
    \resizebox{\textwidth}{!}{
    \begin{tabular}{m{1.7cm}||m{2.8cm}m{3cm}m{3cm}m{4.5cm}}
    
    \toprule
    \textbf{Ref.} & \textbf{Strategy} & \textbf{Horizon} & \textbf{Dataset$/$Availability} & \textbf{Limitations}
    \\ \midrule
    
    Ding~\etal~\cite{ding2014violence}  & 3D convolution with back propagation strategy & Sports violence, applicable to surveillance & Hockey fight$/\checkmark$ & Deep 3D model to directly process raw inputs for features extraction without any prior knowledge. \\ \midrule
    
    Meng~\etal~\cite{meng2017trajectory}  & Integrating frame of trajectory and Deep CNN to detecting human violent behavior in videos & Surveillance and movies & Hockey Fights and Crowd Violence dataset$/\checkmark$ &  Utilizing hand-crafted and deep features in a unified framework, yielding supremacy over SOTA using Hockey Fight and Crowd Violence dataset but lower generalization potentials. \\ \midrule
    
    Mu~\etal~\cite{mu2016violent}  & CNN are used to detect violent detection based on acoustic information & Movies & MediaEval 2015$/\times$&  This framework has limited potentials for real-world surveillance due to the modalities used as input. \\ \midrule
    
    Xia~\etal~\cite{xia2018real}  & Bi-channels CNN with SVM are used for violence detection & Surveillance & Hockey fight, violent flow$/\checkmark$ & Real-time performance (30 (fps) using appearance and motion features, fed to SVM for classification, where SVM can be replaced with more robust sequential deep model.  \\ \midrule
    
    Serrano~\etal~\cite{serrano2018fight}  & Proposed hybrid features based framework for violent behavior & Surveillance and movies & Hockey, Movies and Behave dataset$/\checkmark$ & Hybrid (handcrafted$/$learned features) with computational efficiency for VD task. The authors oppose the idea of only using motion features directly. \\ \midrule
    
    Ullah~\etal~\cite{ullah2019violence} & 3D CNNs with Softmax classifier for VD & Surveillance & Hockey fight, violence in movies, and violent crowd datasets.$/\checkmark$ & Almost real-time method, has higher accuracy but not tested on surveillance-specific data. \\ \midrule
    
    Li~\etal~\cite{li2019efficient} & Proposed 3D convolutional neural networks exclusively for encoding temporal information & Surveillance and movies & Hockey Fights, Movies Dataset and and VIolent-Flows Dataset $/\checkmark$ & The internal design comprises bottleneck units for motion patterns with DenseNet structure for channel's interaction, thereby capturing spatial and temporal information effectively but highly computational. \\ \midrule
    
    Traore~\etal~\cite{traore2020violence}  & RNNs with 2D CNN for VD & Surveillance & Hockey fight, violent flow$/\checkmark$ \newline Real-life violence situations datasets$/\times$ & Optical flow along with RNN for temporal information and 2D CNN for spatial information produces effective results, where the authors achieved best results over three benchmark datasets yet the framework is not tested on real-world videos. \\ \midrule
    \end{tabular}
    }
    \label{tab:DLforVD}
\end{table*}

\subsubsection{Spatial deep models for VD}
Spatial deep models are those utilizing frame-level features to classify activities into violent or non-violent. Spatial methods using deep learning are rare, as VD patterns occur in sequence of frames, therefore, most of the existing research is based on sequential features.

Working with spatial data is not always effective for VD task~\cite{khan2019cover}, as proved from experimental results of Sumon \etal~\cite{sumon2020violence}. The authors experimented on various deep learning models including VGG-19 and ResNet using transfer learning strategy to extract features from the last layer followed by fully connected layer to from a VD classification model. Their results reported significant training accuracy but the testing phase indicated the poor generalization performance of such models, questioning their practical applicability.

\subsubsection{Deep sequence learning for VD}
Deep sequence learning approaches are comparatively robust towards diverse type of environments, holding representative features extraction potentials~\cite{accattoli2020violence}. One of the main reasons of their better generalization abilities is that violent activities occurs in sequence of frames, and these techniques process a single sequence while classifying violent activity. Thus, better representation is achieved for each sequence due to abrupt patterns easily recognizable by deep model when compared with the normal sequence of frames.

Working in sequence learning direction, the existing literature has either considered deep features of a sequence of frames with LSTM~\cite{abdali2019robust} or employed 3D CNNs~\cite{ullah2019violence} to process a sequence of predefined number of frames. For instance, a former research has considered~\cite{sudhakaran2017learning} frame-level features followed by an LSTM with convolutional gates for aggregation. The authors claimed to represent motion patterns effectively using this strategy, that being proved from experimental results, where they received convincing accuracy. LSTM is widely used in such cases of sequence classification~\cite{hussain2019cloud} in many computer vision tasks, where the preprocessing features varies from method to method. For instance, Fenil~\etal extracted local HOG features and fed them to a bi-directional LSTM to retrieve information in both forward and reverse directions~\cite{fenil2019real}. 

\color{black}
The current achievements in VD domain were discussed in this section, where it is evident from existing literature that mainstream state-of-the-art techniques are based on deep learning models. Currently, the spatio-temporal methods perform best in classifying violent activities, otherwise the 3D ConvNets-based methods have higher number of parameters yet lowered performance. The 3D ConvNets mainly focus on the temporal information and learning but limited spatial representation reduces their performance.

\color{black}
\section{Datasets and Performance Evaluation of VD Techniques}
\begin{figure}
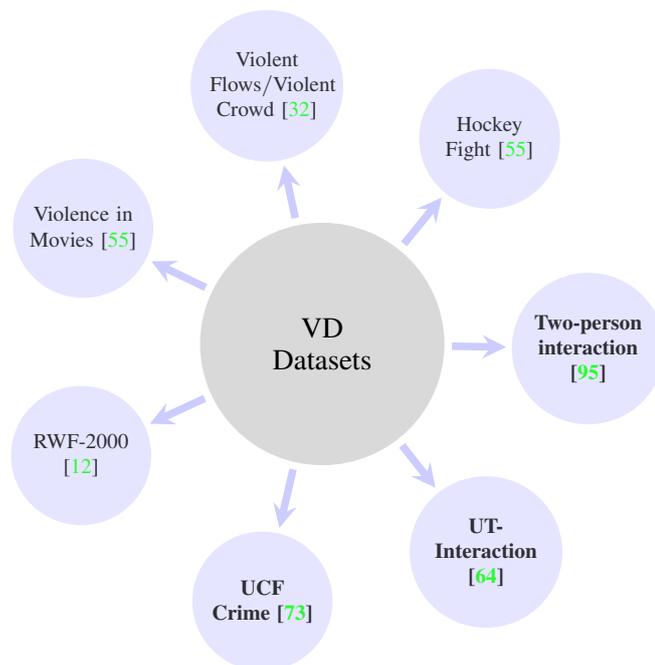

    \centering
    
    \smartdiagram[constellation diagram]{VD \\ Datasets, Hockey Fight \cite{nievas2011violence},  Violent Flows$/$Violent Crowd \cite{hassner2012violent}, Violence in \\ Movies \cite{nievas2011violence}, RWF-2000 \cite{cheng2021rwf}, \textbf{UCF Crime \cite{sultani2018real}}, \textbf{UT-Interaction \cite{ryoo09}}, \textbf{Two-person interaction \cite{yun2012two}}} 
    
    \caption{The available VD datasets. Datasets written in bold letters indicate that these are not purely VD datsets, but contain some classes with human violence and can be utilized in VD domain.}
    \label{fig:VDDatasets}
    
\end{figure}

The datasets particularly relating VD domain are very scarce in existing state-of-the-art literature. Mainstream datasets are either extracted from several sports or movies, hardly resembling real-world surveillance scenarios. Although the patterns relate fight, but still the camera motion and other similar parameters affect a model's performance whenever employed for real-world monitoring. The most challenging datasets so far in VD domain is Violent-Flows and Real-World Fighting (RWF), that contain real-world video footage of violence in crowd, recorded using surveillance low-resolution cameras. The overall datasets are given in Figure \ref{fig:VDDatasets}.

\subsection{Violence in Movies}
As indicated by the name, this dataset is retrieved from movies with fight scenes and actions \cite{nievas2011violence}. This is comparatively smaller in volume, containing only 200 video clips of size 360 by 250 pixels. The first-person motion in this dataset is negligible or even no motion at all in some scenarios, making it very easy to detect the patterns with abnormal behavior i.e., the ones with fight.

\subsection{Violent Crowd$/$Violent Flows}
This dataset is a challenging one as it comprises several categories of violence such as violence in sports and in crowds \cite{hassner2012violent}. The videos are downloaded from YouTube and the overall dataset consists of five sets of video clips, while each one has two classes i.e., violent and non-violent. This dataset is comparatively challenging against other datasets due to data diversity and the reduced amount of videos present in this dataset.

\subsection{Hockey Fight}
This is a very common and widely used dataset for violence detection by most of the methods in VD related literature. The dataset was proposed by Nievas \etal \cite{nievas2011violence}, containing 1000 short video clips. The videos presented in this dataset were previously record in National Hockey League. The dataset is evenly balanced with 500 videos categorized as fight and the remaining 500 videos are placed in non-fight category i.e., normal videos. Each of the video clip contains round about 50 frames and the resolution of these videos is constant, that is 360 by 288. As mentioned, the dataset is recorded in hockey ground, indicating the nature of videos as indoor. This dataset is not much challenging as the videos have very simple activities patterns, therefore, most of the recent methods with deep learning strategies have achieved round about 98\% accuracies on this dataset.

\begin{figure}
    \centering
    \begin{tikzpicture}
        \begin{axis}
        [ybar,
        width=0.9\columnwidth,
        ylabel={\ Accuracy (\%)},
        xlabel={\ Methods},
        enlarge x limits = 0.1,
        xtick=data,
        x tick label style={rotate=45,anchor=east},
        nodes near coords,
        nodes near coords align={vertical},
        symbolic x coords=
        {
        Hassner et al.,
        Ding et al.,
        Bilinski et al.,
        Mabrouk et al.,
        Sudhakaran et al.,
        Xia et al., 
        Ullah et al. (a),
        Li et al.,
        Lohithashva et al.,
        Traoré et al., 
        Ullah et al. (b),
        Ullah et al. (c),
        Freire et al.,
        },
        nodes near coords,
        every node near coord/.append style={rotate=90, anchor=west}
        ]
        \addplot [draw=black, line width=.2mm, semithick, pattern = north east lines, pattern color = black] coordinates
        {
        (Hassner et al., 58.2)
        (Ding et al., 91.0)
        (Bilinski et al., 93.4)
        (Mabrouk et al., 88.6)
        (Sudhakaran et al., 97.1)
        (Xia et al., 95.9)
        (Ullah et al. (a), 96.0)
        (Li et al., 98.3)
        (Lohithashva et al., 91.5) 
        (Traoré et al., 96.5)
        (Ullah et al. (b), 98.5)
        (Ullah et al. (c), 98.2)
        (Freire et al., 99.4)
        };
        
        \end{axis}
    \end{tikzpicture}
    \caption{Performance evaluation of representative VD methods over Hockey fight dataset. The methods are sorted year-wise, starting left-side from 2012 and so on. The references are given as: Hassner et al.~\cite{hassner2012violent}, 
    Ding et al.~\cite{ding2014violence},
    Bilinski et al.~\cite{bilinski2016human},
    Mabrouk et al.~\cite{mabrouk2017spatio},
    Sudhakaran et al.~\cite{sudhakaran2017learning},
    Xia et al.~\cite{xia2018real}, 
    Ullah et al. (a)~\cite{ullah2019violence},
    Li et al.~\cite{li2019efficient},
    Lohithashva et al.~\cite{lohithashva2020violent},
    Traoré et al.~\cite{traore2020violence}, Ullah et al. (b)~\cite{ullah2021ai}, Ullah et al. (c)~\cite{ullah2021intelligent}, Freire~\etal~\cite{freire2022inflated}.}
    \label{fig:hockeyfightevaluation}
\end{figure}
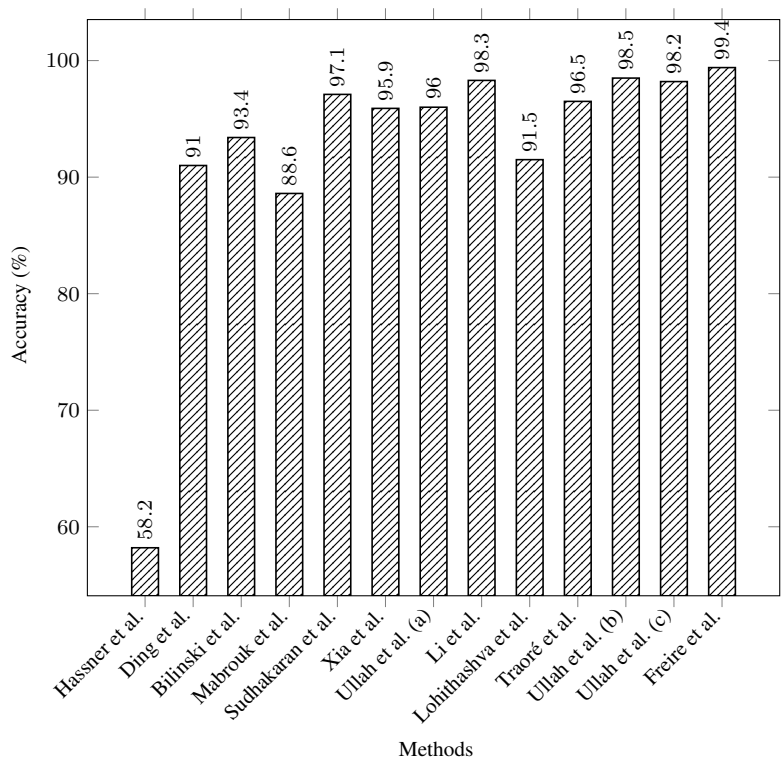

The accuracy of representative VD methods over Hockey fight dataset are given in Figure \ref{fig:hockeyfightevaluation}. Since the nature of the dataset is very simple \ie binary classification problem, therefore, many methods have scored significantly higher accuracy.


\subsection{RWF-2000}
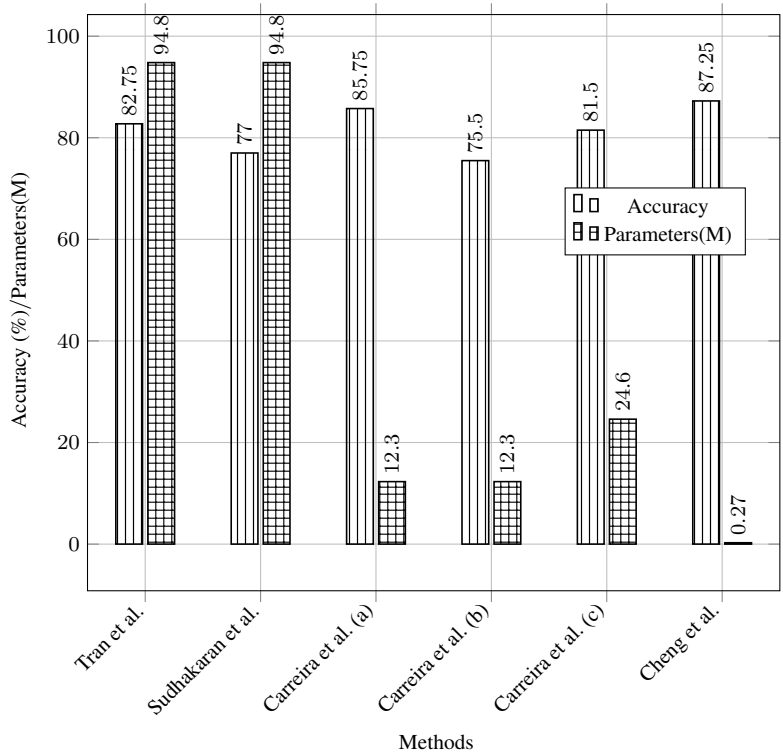
\begin{figure}
    \centering
    \begin{tikzpicture}
        \begin{axis} 
        [ybar,
        width=0.9\columnwidth,
        ymajorgrids=true,
        xmajorgrids=true,
        ylabel={\ Accuracy (\%)$/$Parameters(M)},
        xlabel={\ Methods},
        enlarge x limits = 0.1,
        xtick=data,
        legend style={at={(0.82,0.70)},anchor=north},
        x tick label style={rotate=45,anchor=east},
        symbolic x coords=
        {
        Tran et al.,
        Sudhakaran et al.,
        Carreira et al. (a),
        Carreira et al. (b),
        Carreira et al. (c),
        Cheng et al.
        },,
        nodes near coords,
        every node near coord/.append style={rotate=90, anchor=west}
        ]
        
        \addplot 
        [draw=black, line width=.2mm, semithick, pattern = vertical lines, pattern color = black]
        coordinates 
        {
        (Tran et al.,82.75)
        (Sudhakaran et al.,77.0)
        (Carreira et al. (a),85.75)
        (Carreira et al. (b),75.50)
        (Carreira et al. (c),81.50)
        (Cheng et al.,87.25)
        };  
        
        \addplot 
        [draw=black, line width=.2mm, semithick, pattern = grid, pattern color = black]
        coordinates 
        {
        (Tran et al.,94.8)
        (Sudhakaran et al.,94.8)
        (Carreira et al. (a),12.3)
        (Carreira et al. (b),12.3)
        (Carreira et al. (c),24.6)
        (Cheng et al.,0.27)
        }; 
        \legend  {Accuracy,Parameters(M)}
        
    \end{axis}
    \end{tikzpicture}

    \caption{Performance evaluation of representative VD methods over RWF-2000 dataset, where results are reported based on findings of a prior research \cite{cheng2021rwf}. Carreira~\etal \cite{carreira2017quo} refers to I3D features-based VD, where (a) refers to only RGB data, (b) indicates only optical flow, and (c) represents two stream deep model \ie combining RGB and optical flow. The research methods are presented by Tran~\etal~\cite{tran2015learning},
    Sudhakaran~\etal~\cite{sudhakaran2017learning},
    Carreira~\etal~\cite{carreira2017quo}, and
    Cheng~\etal~\cite{cheng2021rwf}, respectively.}
    \label{fig:rwfevaluation}
\end{figure}

The contents of mainstream existing VD datasets are prepared in some controlled environments, where the quality and visibility of the actors are distinct.  However, violence activities in real-world surveillance pose various resolution, small scale, and different viewpoints. Therefore, to make the violence more authentic and realistic RWF dataset is prepared from the YouTube repository, particularly, surveillance CCTV cameras. It consists of 2000 video clips with five seconds duration and variable resolutions.  It has two classes \ie violent and non-violent activity. This dataset is very challenging because some videos are in poor imaging quality due to dark environment, fast moving actors, and lighting blur. Performance evaluation of representative methods such as ConvLSTM~\cite{tran2015learning}, C3D model~\cite{sudhakaran2017learning}, and I3D architecture~\cite{carreira2017quo} is given in Figure~\ref{fig:rwfevaluation}.

\color{black}
\subsection{Results Discussion}
VD approaches in the early days depended on low-level hand-crafted features such as appearance, histograms of oriented gradients, and different types of motion measurement techniques, such as optical flow. In contrast, complex violent patterns that deviate from normal motion are quite hard to detect using these low-level features. That is why traditional low-level features-based techniques are comparatively less effective on simple datasets such as Hockey fight, which have conventional low-level features-based techniques. The motion patterns in Hockey fight dataset are not very complex, yet the best traditional features-based methods such MoSIFT+HIK~\cite{nievas2011violence}, ViF~\cite{hassner2012violent}, and MoSIFT+KDE+Sparse Coding~\cite{xu2014violent} techniques scored 90.9\%, 82.9\%, and 94.3\% accuracy, respectively. Similarly, on Violent-Flows dataset, the best hand-crafted features methods~\ie ViF~\cite{hassner2012violent}, MoSIFT+KDE+Sparse Coding~\cite{xu2014violent}, and ViF+OViF~\cite{gao2016violence} achieved 81.3\%, 89.05\%, and 88\% accuracy, respectively. Although the mentioned accuracy and performance is higher, but still the deployment of these methods in real-world environments is questionable. As the real-world actions and events are complex as well as their patterns are significantly varying from those of the Hockey fight dataset.

In contrast, the performance of VD techniques emerged significantly with the usage of deep sequential learning algorithms. The emerging recurrent neural networks~\cite{zaremba2014recurrent} are widely adopted in many video classification tasks~\cite{ullah2021conflux,ramanujam2021human} due to their long-range accurate temporal learning abilities. The fact that temporal dependencies as well as spatial information play an important role in sequential patterns classification is evident from Figure~\ref{fig:hockeyfightevaluation} and~\ref{fig:rwfevaluation}. These figures show that the best performance is achieved by deep models containing spatial and temporal learning potentials. For instance, Ullah~\etal~\cite{ullah2021ai} proposed convolutional LSTM followed by GRU and fully connected layers to classify violent and non-violent classes. The authors' motivation of choosing convolutional LSTM with GRU is to  generate effective spatial representation of each frame in video using CNNs and encode the temporal changes and learn long-range dependencies using recurrent neural networks.

In terms of time complexity and the number of parameters, mainstream deep learning-based VD models share limited information~\cite{ullah2019violence} in their research, but the architecture mostly suggest very high number of parameters. For instance, multi-stream convolutional and optical flow networks operations along with multi-layers of LSTMs are highly computational but their details are skipped in the relevant research article~\cite{ullah2021intelligent}. VD experts in computer vision domain mostly focus on the accurate predictions of their method, rather pointing out the real-time decision making of a technique. The number of parameters for different deep models can be analyzed from Figure~\ref{fig:rwfevaluation}, where most of the deep models focus on the accuracy of their method. The number of parameters are mostly high, except some representative methods that are able to provide a balanced trade-off between the number of parameters and the accuracy of the model. For instance, Carreira~\etal~\cite{carreira2017quo} I3D features-based mechanism to compute VD has 12.3 million parameters when computed using only RGB or optical flow input, while their two stream model has two times parameters (24.6 million) compared to the individual streams. The best single stream accuracy is 85.75\% for RGB input, while the two stream architecture's accuracy is 81.5 and the optical flow only mechanism produced the lowest 75.5\% accuracy. Although these methods have comparatively lower parameters against state-of-the-art, as given in Figure~\ref{fig:rwfevaluation}. The best achievement and trade-off between parameters and accuracy is provided by Cheng~\etal~\cite{cheng2021rwf}, where the authors achieve 87.25\% accuracy with only 0.27 million parameters. They achieved reduced parameters by employing the concept of depth-wise separable convolutions from  Pseudo-3D Residual Networks~\cite{qiu2017learning} and MobileNet~\cite{howard2017mobilenets} to adopt the 3D convolutional layers in their proposed model, which tremendously reduced their model parameters without compromising on the performance loss.

\color{black}
\section{Challenges and Future Directions in VD Domain}
Action and activity recognition is one of the most interesting and hot area of research, covering several major domains. One of these major domains is VD, that seems not to be fully covered to the level of its requirements in today's smart surveillance systems. For instance, there have been extremely challenging datasets available for action and activity recognition, while in contrast, there are only few for VD domain. Similarly, there are many interesting breakthroughs in the field of activity recognition such as future activity perceptions, activities localization, multi-view action and activity recognition. VD domain is being deprived of such breakthroughs, as it lacks multi-view challenging datasets recorded in real-world scenarios such as UCF anomaly detection~\cite{roka2022review} and recognition dataset \cite{sultani2018real}. Similarly, the accuracy of the current methods is so perfect on the available datasets but these data does not seem to be representative of real-world because these data are retrieved or recorded from sports videos. The major challenges in VD domain are covered in the subsequent paragraphs along with possible future directions.

\subsection{Scarcity of Challenging VD Data}
The video data available for use in VD methods is very limited \ie only few datasets are there to be used for comparison with SOTA methods. Practical applicability of action and activity recognition methods highly depends on the used data while training. For instance, diverse set of videos while training a deep model or learning motion patterns of huge set of violent videos results in highly generic trained weights or a mathematical model for practical usage in real-world scenarios.

\subsection{Evaluation of VD Methods}
So far, confusion matrix, accuracy, and F1 metrics are commonly used to evaluate VD model or compare its performance against SOTA methods. Current VD trend is only focused on accuracy of any model, ignoring the worth of time complexity for resource-restricted environments. Furthermore, the evaluation is mostly performed using test set of the same dataset, that is a biased decision until the test set is marginally different from the training set, evaluating a model's generalization.

\subsection{Future VD Prediction}
Future activity prediction has gained certain importance recently with many applications to autonomous driving for human location and activity perception. The idea of future activities prediction is also required in VD domain, predicting future violent actions with significant accuracy will lead to prevention of the receipted action or at least quick responsive actions initiation.

\subsection{Generalization Potentials of VD Methods}
Generalization abilities of any VD method has direct relation with its applicability, the more is a model generic and performs best towards unseen data, the greater are the chances to detect violence in early stages. Such type of task can be achieved by specially designing a neural network to function completely or consider the importance of certain parameters including human locations, their distance among each other, and their motion trajectories. Such parameters have importance to detect violence and helps in distinguishing abnormal violence patterns form real-time normal video frames. Existing research on anomaly detection by Ullah \etal can be a good source to integrate various importance features and gain enhanced and generalized model \cite{ullah2021cnn}. Similar research is applied to activity recognition domain \cite{ullah2018activity}.

\subsection{Relating Anomaly Detection with VD}
Anomaly detection domain has broader number of classes such as fire, arson, shoplifting, and others including human violence, comprising assault and fighting as well \cite{ullah2021efficient}. So far, VD literature is based on just binary classification problem \ie detecting whether there is a human fight or not, which is a simple problem for today's computer vision algorithms. Complex problems such as further classifying the type of fight and its intensity level are future directions yet to be covered, demanding experts' attention. In this regard, end-to-end deep learning models can be employed to classify any abnormal fight action and relate its intensity by using the same models' parameters.

\subsection{Edge-based VD Methods}
To ensure the safety of people, eradicate or reduce the risk of the violence, it is necessary to equip the vision sensors in smart cities with highly computational embedded devices. Besides the embedded devices, the vision sensors themselves can be made smart enough to process video data effectively for any kind of prior violence acts detection or violence detection while it's happening.

It is very challenging task to achieve effective VD over the edge networks due to higher computational cost of mainstream deep learning CNN \cite{suleiman2017towards} models \cite{lygouras2019unsupervised}. Furthermore, online learning over the edge devices is also a challenging task as the traditional training is performed over high-computation devices. Future directions in terms of VD is to acquire optimized online learning techniques that are able to learn over the edge devices \cite{cui2019stochastic,jin2021learning}.

\subsection{Reinforcement Learning for VD}
Reinforcement learning is widely been used in many computer vision problems \cite{dogru2021actor,hafiz2021reinforcement}, where it has not been explored for the problem of VD in real-world environments. Reinforcement learning techniques have surely accomplished complex tasks with higher effective and efficient results. The current limited adaptability with the environment challenge for VD methods in real-world can be easily solved with the assistance of reinforcement learning methods \cite{agarwal2021contrastive,sonar2021invariant}.

\subsection{Vision Transformers in VD Domain}
Vision transformers in sequence-based problems have shown tremendous performance, particularly, for image recognition and detection tasks \cite{dosovitskiy2020image,zhang2021vit}. Similarly, TimeSformer are introduced for precise video classification tasks such as action and activity recognition and video understanding \cite{bertasius2021space}. There are many other recent technologies such as human brain-inspired spiking neural networks \cite{pavlidis2005spiking}, and explainable artificial intelligence \cite{arrieta2020explainable,rojat2021explainable} models, that are barely explored for the VD task. These possible future directions need the research attention in near future to enrich this domain and assist significantly in automated surveillance, thereby protecting and ensuring the safety of humans.

\subsection{\textcolor{black}{Computation Friendly VD Methods}}
\textcolor{black}{The computation resources acquired by the current VD methods are not environment friendly and most of the techniques utilize computationally complex models to detect violent activities. Such methods practical applicability is questionable as in most of the surveillance environments real-time decision making is required for quick responsive actions and counter-measures. Therefore, future research from computation perspective with end-to-end resources friendly deep models can be applied in VD domain.}

\section{Conclusive Remarks}
The presented review paper highlights the current achievements of computer vision experts in VD domain and highlights the major challenges with future research directions. Concluding the overall review, it suggests the in the early VD research, traditional features such as motion vectors, acceleration information, instant change detection filters are utilized to identify violence. With the success of deep learning in many domains, the traditional processing is replaced by the powerful features representation potentials of 2D CNNs, that too comes with some limitations such as non-effective complex large motion patterns representation, sometimes huge time complexity, under representation of events, and the foremost problem of thirst of data. VD literature advanced to a new level with the usage of 3D CNNs for video data representation, which consider time series frames for video classification tasks and performs well for VD, but with the huge cost of computation, requiring complex GPUs for output generation.

\color{black}
There is a higher level of accuracy and precision in the testing results for most of the VD datasets surveyed in this study. However, the existing methods do not focus enough on the training and evaluation of their models in terms of the environment that they will live in in the real world. It can for instance happen that in real world surveillance scenarios, a scene may contain multiple abrupt changes, such as the sudden appearance of irrelevant objects that instantly change the motion patterns, and hence destabilize the model's output prediction. Thus, there is a need to adapt deep models in terms of non-stationary surveillance environments in order to cope with the current situation. Furthermore, it is also necessary to use VD techniques focusing on meaningful information extraction. This is to analyze the person's behavior after its successful localization has been accomplished, finally resulting in successful VD and localization, respectively.
\color{black}


%
\section*{Conflict of Interest}

The authors declare that they have no known conflict of interest.

\bibliographystyle{spmpsci}      
\bibliography{Main}   

\end{document}